\documentclass[11pt]{article}
\usepackage[letterpaper,margin=0.85in]{geometry}
\usepackage{amsmath,amssymb,booktabs,graphicx,array,tabularx,microtype}
\usepackage[colorlinks=true,allcolors=blue]{hyperref}
\usepackage[font=small,labelfont=bf]{caption}
\usepackage{enumitem,needspace}
\setlist[itemize]{leftmargin=*,nosep}
\newcommand{\fittraceurl}{https://github.com/SeverinVisionary/prior-shift-diagnostic/blob/a5c1bdc45a69c496d37d4a482fff64f73b537046/data/g3_qwen7b/Qwen2.5-7B-Instruct@a09a35458c70/result.json}
\title{Less Loss, Fixed Weights: Logit Corrections for Forgetting}
\hypersetup{pdftitle={Less Loss, Fixed Weights: Logit Corrections for Forgetting},pdfauthor={Hanyu Yang}}
\author{Hanyu Yang\\\small ORCID: \href{https://orcid.org/0009-0005-0419-4070}{0009-0005-0419-4070}}
\date{October 2026}
\begin{document}
\maketitle
\begin{abstract}
When a language model is fine-tuned to follow instructions, or further pretrained on a new subject, it often gets worse at predicting standard web text. This is known as “forgetting.” We ask whether this loss can be undone by simple, fixed adjustments to the model’s final token predictions. Across 14 pairs of base and derived models, from 1B to 7B parameters, we fitted a fixed bonus or penalty for each token (a vocabulary bias), alone or together with a temperature.

On text held out from correction fitting, the largest top-4K bias recovery estimate is under 10\%, with loss conditional on the retained vocabulary. Combining the bias with a temperature recovers up to 36\%. Models further pretrained on math lose the most and recover almost nothing. Finally, in a controlled simulation where a model is damaged at random, with no planted bias or temperature change, the bias-plus-temperature fix still recovers 29\%. Most of the real-model loss increase stays out of reach of our restricted fits.
\end{abstract}

\section{Introduction}
When AI developers take a fully trained language model and fine-tune it to follow instructions, or train it further on math or code, the model frequently “forgets” some of what it learned before. It then predicts the broad, unstructured web text its base model handled well less accurately. We ask how much of that loss a fixed adjustment to its final predictions can recover.

If post-training forgetting were mostly caused by the model developing an artificial bias for or against certain tokens, we could reverse it with a single, simple adjustment at the very end of its prediction process, without retraining the network.

Recent studies hint that forgetting and output bias might be linked. Zhu et al. found that forgetting in image-and-text models moves together with an output-token bias \cite{zhu}. Hu et al. showed that alignment makes models far more overconfident on benchmarks \cite{hu}. Ren et al. distinguish collision and erosion; on multiple-choice tasks, they track probability draining away from the answer letters \cite{ren}. Logit-bias corrections already serve multiple-choice debiasing \cite{pride} and black-box adaptation \cite{cohen}. Related recovery methods use conjugate prompting \cite{kotha} or shared shifts and update directions \cite{palit}. Our question concerns recovery of web-text loss.

Measuring this is tricky. If you tailor an adjustment to a specific text, it can make any model look better, including the original base model. If we don’t account for this, we overestimate how much was recovered. To be fair, we fit the same kind of fix to both the original and the derived model, and we report the extra benefit the derived model receives. We call this the “symmetric recovered share.”

We measured this share on 14 base/derived pairs, scoring one million tokens of standard web text. To check that our numbers weren’t misleading, we also ran simulated tests where we know the true cause of the damage.

\section{Measurement}
\paragraph{Text and logits.}
We use documents from one DCLM shard in the OLMo-2 pretraining mix. For OLMo-2, this probably matches its training distribution. SmolLM2 also lists DCLM among its sources. Exposure of Qwen2.5 to these documents is unknown. For each tokenizer, we truncate documents to 2,047 tokens and prefix them with BOS, or EOS when BOS is absent. We assign documents in file order to FIT and then EVAL until each split contains at least one million text tokens before target masking. The first 90\% of FIT documents fit the corrections; the last 10\% select them. EVAL stays separate.

We feed both models the base tokenizer's token IDs. Tokenizer IDs are compared on one probe sentence. From cached bf16 hidden states and output weights, we reconstruct logits with fp32 accumulation. We mask special and added IDs across each base-model group's tokenizers, plus output rows beyond their smallest vocabulary, leaving $V_{\rm reg}$. Masked targets are excluded. All losses are therefore conditional on this vocabulary. Loss from probability assigned to excluded tokens is outside this measurement.

\paragraph{Corrections.}
For allowed token $v$ and context $x$, the combined correction is
\begin{equation}
 p_{b,T}(v\mid x)=\frac{\exp(z_v(x)/T+b_v)}{\sum_{u\in V_{\rm reg}}\exp(z_u(x)/T+b_u)}.
\end{equation}
Bias alone fixes $T=1$. We fit each model separately. The bias uses iterative proportional fitting toward smoothed FIT counts, following marginal adaptation \cite{kneser}. The combined form is bias-corrected temperature scaling \cite{alexandari,guo}; our fitting procedure differs from joint maximum likelihood.

The target distribution is $t=(c+0.1N\bar p)/(1.1N)$, where $c$ counts FIT training targets, $N=\sum_v c_v$, and $\bar p$ is the model's uncorrected mean prediction on them. This target stays fixed across temperatures. Each bias step adds $\log t-\log\bar p_b$ on active coordinates. We retain the lowest validation-loss iterate, including $b=0$, with at most 20 steps and patience 3. Full-vocabulary bias-only fits selected $b=0$ for every released model. We therefore restrict active biases to the 4,096 most frequent training targets; all other biases remain zero. The softmax still includes $V_{\rm reg}$.

For bias plus temperature, 14 golden-section evaluations in $\log T$ search $T\in[0.25,4]$. Each restarts the bias at zero; the lowest validation loss selects both the iterate and temperature. The search does not explicitly include $T=1$ or certify a global optimum. Qwen2.5 7B Instruct's top-4K bias-only fit reached the cap with validation loss still falling (\href{\fittraceurl}{trace}).

\paragraph{Recovered share.}
Let $L_m$ be token-weighted EVAL cross-entropy for model $m$, and let $L_m^c$ use its own fitted correction $c$. Set $\Delta=L_{\rm derived}-L_{\rm base}$. We report
\begin{equation}\label{eq:share}
 S_{\rm sym}^c=\frac{(L_{\rm derived}-L_{\rm derived}^c)-(L_{\rm base}-L_{\rm base}^c)}{\Delta}
 =1-\frac{L_{\rm derived}^c-L_{\rm base}^c}{\Delta}.
\end{equation}
The derived-only share is $S_{\rm raw}^c=(L_{\rm derived}-L_{\rm derived}^c)/\Delta$. Shares are undefined when $\Delta=0$ and are not clipped to $[0,1]$. For $\Delta>0$, a negative symmetric share means the base gain is larger; a positive share can also occur when both corrected losses rise but the base loss rises more. Small symmetric shares can mask gains shared by both models.

We use paired document bootstrap intervals: 2,000 resamples of EVAL documents with replacement, retaining each document's token count and both models' losses. Corrections stay fixed. These intervals cover EVAL sampling variation, with fitting variability excluded. Pairs sharing a base model and documents are dependent.

\begin{figure}[tb]
\centering\includegraphics[width=0.95\linewidth]{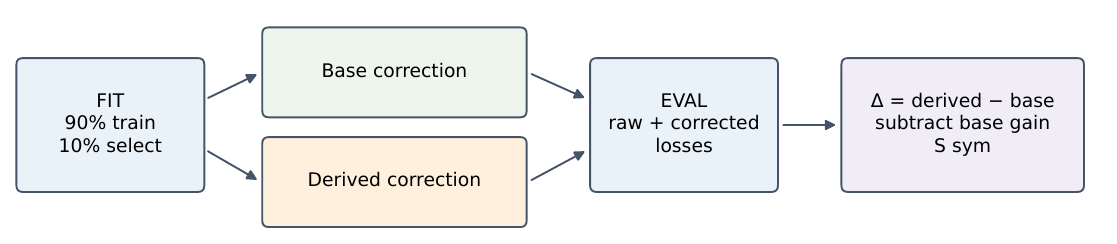}
\caption{FIT selects a separate correction for each model. EVAL supplies the uncorrected gap $\Delta$ and both gains used in Equation~\ref{eq:share}.}
\label{fig:pipeline}
\end{figure}

\section{Results}
\begin{figure}[p]
\centering\includegraphics[width=\linewidth]{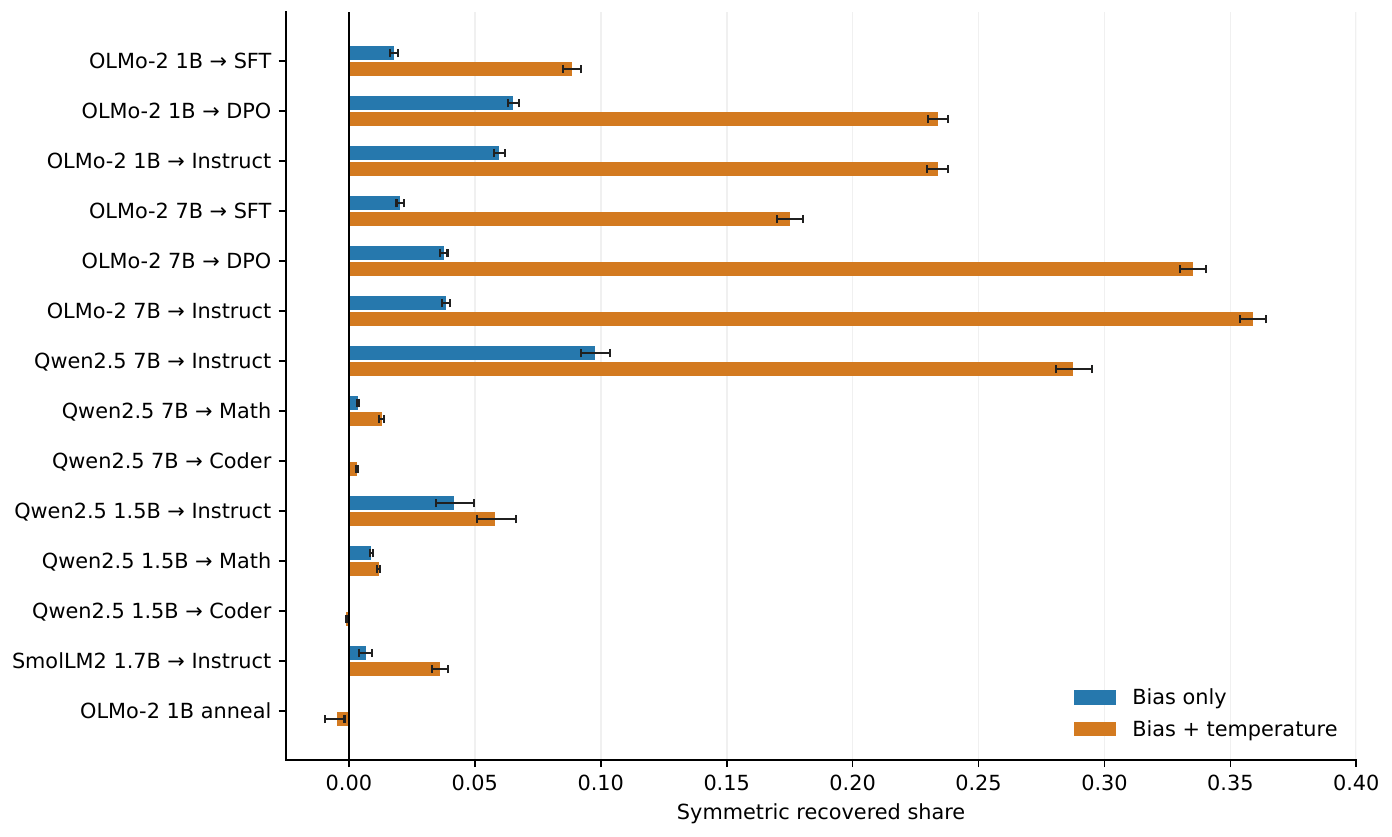}
\caption{Symmetric recovered shares for the top-4K bias and combined fits, with 95\% paired document-bootstrap intervals. Negative estimates are retained. Fits are held fixed in the bootstrap.}
\label{fig:shares}
\end{figure}

\paragraph{Bias-only recovery is small under this fit.}
The largest top-4K bias point estimate is 9.8\%, for Qwen2.5 7B $\to$ Instruct (95\% interval: 9.2\%--10.4\%); every other point estimate is below 7\% (Figure~\ref{fig:shares}). For Qwen2.5 7B $\to$ Instruct, an EVAL-fitted one-step log-ratio correction assigns zero probability to tokens absent from EVAL and reaches 15.8\% symmetric recovery \cite{kneser}. This in-sample value is neither held-out recovery nor an upper bound on a converged bias fit.

\paragraph{The combined fit helps most on OLMo-2 preference-tuned models.}
Bias plus temperature recovers 35.9\% for OLMo-2 7B $\to$ Instruct and 33.5\% for its DPO checkpoint. At both OLMo-2 sizes, recovery is larger after DPO than after SFT. Qwen2.5 7B $\to$ Instruct reaches 28.8\%, against a small uncorrected gap of 0.050 nats/token. SmolLM2 and Qwen2.5 1.5B $\to$ Instruct stay below 6\%.

\begin{figure}[!b]
\centering\includegraphics[width=\linewidth]{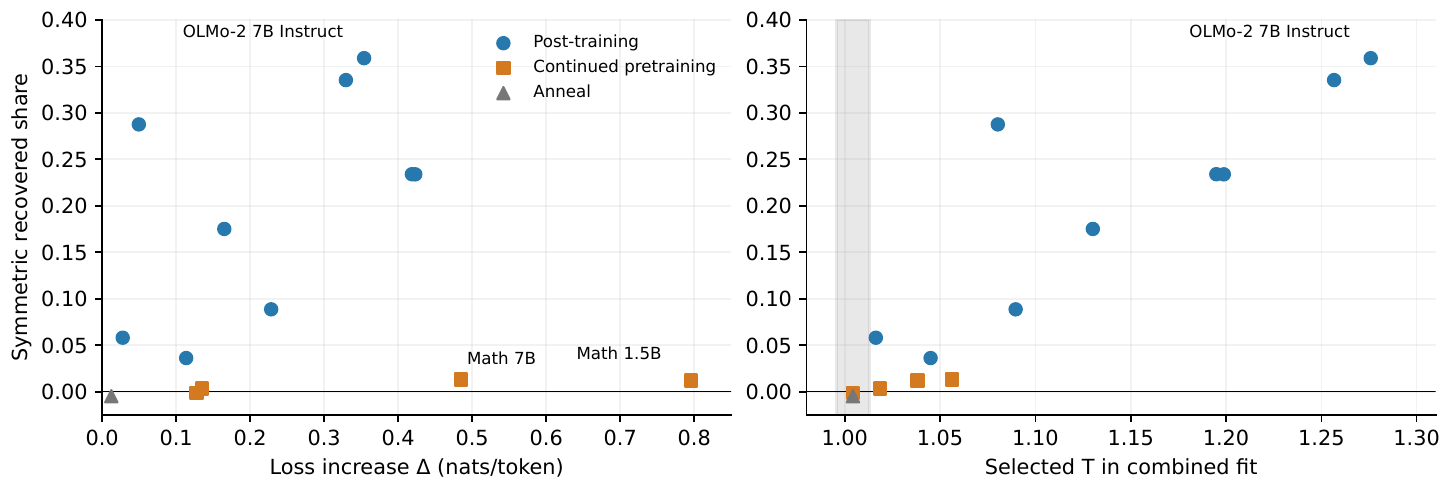}
\caption{Point estimates of symmetric recovery for the top-4K combined fit against the uncorrected loss gap (left) and selected temperature (right). The grey band spans the six base temperatures. Points sharing a base are dependent; no cross-pair trend is fitted.}
\label{fig:scatter}
\end{figure}

The combined fit selects $T=1.08$--$1.28$ for the OLMo-2 post-trained models and Qwen2.5 7B Instruct, compared with $0.996$--$1.013$ for the six bases (Figure~\ref{fig:scatter}). We fitted bias and temperature together, without a temperature-only baseline. These temperatures do not separate temperature shifts from other logit changes.

\paragraph{Math and Coder models recover little.}
The Qwen2.5 Math gaps are 0.485 nats/token at 7B and 0.796 at 1.5B. Their top-4K combined shares are 1.3\% and 1.2\%. Coder gaps are about 0.13 nats/token, with combined shares near zero. Table~\ref{tab:pairs} gives all pairs. Each gap compares models under the same tokenizer; its magnitude across tokenizer groups is descriptive.

\begin{table}[!htbp]
\centering\small
\caption{All 14 pairs. $\Delta$ is in nats/token; shares are $S_{\rm sym}$. Temperatures belong to the top-4K combined fit. SFT, DPO and Instruct rows are post-training; Math and Coder rows are continued pretraining. The anneal row compares intermediate and final pretraining checkpoints.}
\label{tab:pairs}
\begin{tabular}{lrrrr}
\toprule
Pair & $\Delta$ & Bias & Bias + $T$ & $T$ (derived/base)\\
\midrule
OLMo-2 1B $\to$ SFT & 0.228 & 0.018 & 0.089 & 1.090 / 1.004\\
OLMo-2 1B $\to$ DPO & 0.419 & 0.065 & 0.234 & 1.195 / 1.004\\
OLMo-2 1B $\to$ Instruct & 0.423 & 0.060 & 0.234 & 1.199 / 1.004\\
OLMo-2 7B $\to$ SFT & 0.165 & 0.020 & 0.175 & 1.130 / 1.006\\
OLMo-2 7B $\to$ DPO & 0.329 & 0.038 & 0.335 & 1.257 / 1.006\\
OLMo-2 7B $\to$ Instruct & 0.354 & 0.039 & 0.359 & 1.276 / 1.006\\
Qwen2.5 7B $\to$ Instruct & 0.050 & 0.098 & 0.288 & 1.080 / 1.013\\
Qwen2.5 7B $\to$ Math & 0.485 & 0.004 & 0.013 & 1.056 / 1.013\\
Qwen2.5 7B $\to$ Coder & 0.135 & 0.000 & 0.003 & 1.018 / 1.013\\
Qwen2.5 1.5B $\to$ Instruct & 0.028 & 0.042 & 0.058 & 1.016 / 1.010\\
Qwen2.5 1.5B $\to$ Math & 0.796 & 0.009 & 0.012 & 1.038 / 1.010\\
Qwen2.5 1.5B $\to$ Coder & 0.127 & 0.000 & -0.001 & 1.004 / 1.010\\
SmolLM2 1.7B $\to$ Instruct & 0.114 & 0.007 & 0.036 & 1.045 / 1.004\\
OLMo-2 1B anneal & 0.012 & 0.000 & -0.005 & 1.004 / 0.996\\
\bottomrule
\end{tabular}

\end{table}

\clearpage
\section{Synthetic checks and interpretation}
Adding random noise to output weights changes logits differently across contexts. In the saved synthetic test, the symmetric recovered share is 0.290 for the combined fit and 0.076 for bias alone. This is one simulated case at one damage level. No global bias or temperature change was planted. Recovery alone therefore does not identify the training mechanism.

The planted-bias check gives over 90\% derived-only bias recovery and symmetric combined recovery. With doubled logits, derived-only combined recovery exceeds 90\%, and the fitted temperature is 1.987. Identical models give zero gap and undefined shares.

The proxy case keeps the weight noise but shifts the data-generating prior. The base model gains too: subtracting that gain reduces combined recovery from 0.474 (derived-only) to 0.299 (symmetric), as Table~\ref{tab:synthetic} shows.

\begin{table}[htbp]
\centering\small
\caption{Selected saved synthetic results: $V=400$, hidden dimension 48, and 300 FIT/300 EVAL documents of 64 tokens. Gaussian weight noise has standard deviation 0.25; original weights have standard deviation 0.5. Dashes mark unreported shares.}
\label{tab:synthetic}
\begin{tabular}{lrrrr}
\toprule
&\multicolumn{2}{c}{Bias only}&\multicolumn{2}{c}{Bias + temperature}\\
Case&$S_{\rm raw}$&$S_{\rm sym}$&$S_{\rm raw}$&$S_{\rm sym}$\\
\midrule
Random damage&--&0.076&--&0.290\\
Proxy text&0.251&0.076&0.474&0.299\\
\bottomrule
\end{tabular}
\end{table}

Reported synthetic recoveries use full-vocabulary fits, whereas the real-model results use top-4K fits. The 29\% result therefore provides no threshold for the real-model shares. All tokens remain unmasked in the synthetic tests, so the masking code is untested there.

\section{Limits and next measurement}
We cannot determine how much a better context-free fit would recover: vocabulary support, smoothing and stopping rules were fixed, and some fits reached the iteration cap.

We tested one shard of web text, three model families, and models of 7B parameters or fewer. Tokenizer groups can select different documents. This 14-pair comparison is exploratory.

We would next fit temperature alone, then allow the bias fit to converge further while varying its support and the held-out text. This would help distinguish limits of the correction family from limits of fitting or the scoring distribution.

\section*{Serving implementations}
Using these corrections during generation also requires correct sampling. vLLM has a merged fix for reused random draws across block-verification rounds \cite{vllm_rng}. Open vLLM patches address negative CPU draft placeholders \cite{vllm_cpu}, narrower draft vocabularies \cite{vllm_vocab}, bonus-token minimum-length masking \cite{vllm_min}, and repeated thinking-end sequences \cite{vllm_think}. An open SGLang patch removes padded draft siblings from NGRAM sampling \cite{sglang_pad}. Statuses are as of 29 October 2026. We use teacher-forced Transformers/PyTorch losses; neither engine enters these measurements.

\section*{Data and code}
Code and per-document loss sums are at \href{https://github.com/SeverinVisionary/prior-shift-diagnostic}{the public repository}, archived at \href{https://doi.org/10.5281/zenodo.23304208}{doi:10.5281/zenodo.23304208}. This revision uses public commit \texttt{a5c1bdc45a69}. We recomputed the 14 estimates and their bootstrap intervals from the saved sums; model inference and fitting were not rerun. The original sweep used about 13.5 A100-hours. Fitted bias vectors and cached logits were not retained in the public archive.

\section*{Disclosure}
The author chose the question and the models, and checked the results. AI coding and writing assistants (Claude Code, OpenAI Codex, Gemini, ChatGPT) were used to help write and run the code, make the figures and tables, and edit and fact-check the text. The author is responsible for the content.

\end{document}